\documentclass[letterpaper, 10 pt, conference]{ieeeconf}  

\IEEEoverridecommandlockouts                              

\title{\LARGE \bf An Analysis of Human-Robot Information Streams\\to Inform Dynamic Autonomy Allocation}

\author{Christopher X. Miller$^{1}$, Temesgen Gebrekristos$^{1}$, Michael Young$^{1}$, Enid Montague,$^{2}$ and Brenna Argall$^{1,3}$

\thanks{$^{1}$The Department of Mechanical Engineering, Northwestern University, and the Shirley Ryan AbilityLab, Chicago, IL - USA {\tt\small [chris.x.miller, tem, mikesyoung] @u.northwestern.edu}}

\thanks{$^{2}$College of Computing, DePaul University and Feinberg School of Medicine, Northwestern University, Chicago, IL - USA {\tt\small emontag1@cdm.depaul.edu}}

\thanks{$^{3}$Departments of Computer Science and Physical Medicine and Rehabilitation, Northwestern University, Chicago, IL - USA {\tt\small brenna.argall@northwestern.edu}}%

}
\usepackage{enumerate}
\usepackage{color}
\usepackage{graphicx}
\usepackage{subfigure}
\usepackage{multirow}
\usepackage{multicol}
\usepackage{wrapfig}
\usepackage{amsmath}
\usepackage{tabularx}
\usepackage{flushend}

\usepackage{listings}
\begin{document}
\maketitle
\thispagestyle{empty}
\pagestyle{empty}

\begin{abstract}
A dynamic autonomy allocation framework automatically shifts how much control lies with the human versus the robotics autonomy, for example based on factors such as environmental safety or user preference. To investigate the question of which factors should drive dynamic autonomy allocation, we perform a human subject study to collect ground truth data that shifts between levels of autonomy during shared-control robot operation. Information streams from the human, the interaction between the human and the robot, and the environment are analyzed. Machine learning methods---both classical and deep learning---are trained on this data. An analysis of information streams from the human-robot team suggests features which capture the interaction between the human and the robotics autonomy are the most informative in predicting when to shift autonomy levels. Even the addition of data from the environment does little to improve upon this predictive power. The features learned by deep networks, in comparison to the hand-engineered features, prove variable in their ability to represent shift-relevant information. This work demonstrates the classification power of human-only and human-robot interaction information streams for use in the design of shared-control frameworks, and provides insights into the comparative utility of various data streams and methods to extract shift-relevant information from those data.

\end{abstract}

\section{INTRODUCTION}

As robotic partners become ubiquitous in daily life---from self-driving cars on our roads to manufacturing co-robots in our workplaces to caregiving robots in our hospitals---we recognize that control signals issued by human teammates may not always be reliable. A person might require differing levels of support as they become distracted or fatigued, for example due to an adversarial scenario or physical impairment. Within assistive robotics, the ability to adapt to a person's needs are of particular importance as inappropriate amounts of assistance---in particular, overassistance---can result in a loss of independence. An appropriately designed shared-control framework will autonomously adapt its level of support as the human's needs or environments change. A key question to consider, however, is in response to \textit{which factors} should this dynamic allocation of autonomy occur?

Many existing shared-control frameworks are static and do not consider the human partner's changing control signals to allocate control authority. We seek to develop a framework for dynamic autonomy allocation that acknowledges the interactions in the human-robot team are not static. To that end, we present an exploratory study using an autonomous wheelchair where a user can modulate autonomous assistance within a pre-defined, discretized Levels of Autonomy (LOAs) framework. Within this framework, different information streams are recorded and analyzed for their utility to inform autonomy allocation. We cast the problem as one of supervised learning, allowing both the user and the robotics autonomy to request a shift in autonomy level. This work focuses less on the framework and more on determining which \textit{information streams} are most useful for determining when to change the level of autonomous assistance.

The remainder of this paper is organized as follows. In Section II, we review related work within the domains of shared-control and dynamic-autonomy allocation. We present a detailed description of the problem in Section III. Section IV overviews the experimental design. Section V presents the datasets, feature design, and machine learning methods. We present our results in Section VI and discuss our insights for autonomy allocation and future work in Section VII. We conclude in Section VIII. 

\section{BACKGROUND}

The field of \textit{shared control}, seeks to determine the best allocation of control authority between human-robot partners \cite{musicreview}. Within the domain of assistive robotics, systems that share control between the human and the autonomy, rather than allocate all control to the autonomy, are of particular desire and importance \cite{OHorn}. Early shared-control methods optimize for user safety \cite{safetyinsc}. More recently, approaches aim to allow the human as much control as possible. Some methods rely on fixed, user-selectable or adaptable blending schemes to combine the human’s and the autonomy’s commands \cite{linear}. Similarly, to adapt to a human’s specific physical needs, the autonomy can monitor a person's biometric signals \cite{biometricadapt}. Control authority may be derived from recent human-robot team performance as measured by the team’s task outputs \cite{Rahman2015}. Others allocate the autonomy’s control authority based on the human-robot team's state and the autonomy’s inferences about the human’s behavior \cite{prb2}. Within assistive robotics in particular, shared-control schemes may require user personalization \cite{deepak}. 

Levels of Autonomy (LOAs) are a shared-control framework in which a fixed number of discretized LOAs are defined and, within each LOA, the control authority between the human and robot is set according to a predefined paradigm \cite{1978Vedplank}. As the levels increase from lowest to highest, so does the autonomy's control authority. The usefulness of discrete LOAs is demonstrated in studies where a user can request all or no autonomous assistance \cite{LOAsalloc}. The majority of LOA shifting is linear, shifting between adjacent LOAs, though two dimensional spectrums of LOAs have been proposed \cite{LOAsalloc}. Prior work also notes the need to autonomously shift between LOAs \cite{coppinlegras}. 

There are many sources of data to inform the allocation of control within shared-control frameworks. Nearly all shared-control frameworks rely on information gathered from a variety of sensors observing the human-robot team's \textit{environment} \cite{musicreview}, such as cameras \cite{camera} or range finders \cite{taylor}. Others rely on information derived from the \textit{human}, ranging from raw control commands \cite{mahdieharxiv} to unit-less measures of human movement smoothness \cite{SPARC}. More recently, there has been some emphasis on quantifying the \textit{interaction} between the human and the autonomy, for example interaction that compares the autonomy's expected actions to the human's actual actions and adapts accordingly \cite{julieshah}. 

To then use these data streams to inform how control is allocated between the autonomy and human typically relies on some form of manual feature engineering. An alternative however is to process the raw information streams directly \cite{gatechencoder}, requiring little-to-no feature extraction, which also holds the promise of greater generalization across robot platforms. The question of which information streams, and interpretations of them, are most useful to a dynamic autonomy allocation paradigm remains an open research question.  

\section{PROBLEM DESCRIPTION AND IMPLEMENTATION}

In this work, we propose a systematic investigation into which categories of information hold the most utility in signaling when to shift the allocation of control between human and autonomy when sharing control of a robotic device. We identify four categories of information sources within the joint human-robot system (Table \ref{tb:1}): from the Human ($\mathcal{H}$), from the Autonomy ($\mathcal{C}$), from the Human-Robot Interaction ($\mathcal{I}$), and from the External Environment ($\mathcal{E}$).  

\begin{table}[t]
\vspace{0.55em}
\centering
\begin{tabular}{p{3.5cm}|p{4cm}}
\textbf{Information Stream Name} & \textbf{Information Source} \\ \hline 
Human Control Signals ($\mathcal{H}$)  &  Human control inputs \\ \hline
Autonomy Control Signals  ($\mathcal{C}$) & Autonomous planner's control commands \\ \hline
Interaction Signals  ($\mathcal{I}$) & Result of the human-robot interaction (e.g., differences between the human and autonomy control commands); includes the team's task performance \\ \hline
Environment Signals  ($\mathcal{E}$) & Task execution environment \\ 

\end{tabular}
\caption{CATEGORIES OF INFORMATION IN A HUMAN-ROBOT TEAM.}
\label{tb:1}
\vspace{-3em}
\end{table}

 \subsection{Problem Description}
 
 We partition the autonomy, $A$, into $n$ discrete levels $A_i, i \in [0,n]$ where, as $i \rightarrow n$, the control authority shifts from the human to the autonomy. Within this framework, transitions may only occur between adjacent autonomy levels, following $A_i \rightarrow A_{i+1}$ or $A_i \rightarrow A_{i-1}$. 
 
 We frame the problem of autonomy allocation as a question of when to shift between autonomy levels. We cast this as a supervised, multi-class classification problem where a learning algorithm will resolve a function such that $g : X \rightarrow Y$, where $X$ are features and $Y$ are the set of class labels. Here, $Y$ includes three labels: $y_0 =$ \textit{shift-up}, $y_1 =$ \textit{no-shift}, and $y_2 =$ \textit{shift-down}. Feature vectors, $x_i$, are extracted, both manually and automatically (described further in Section V.B), from a subset of the information streams presented in Table \ref{tb:1}. Labels are assigned to each vector according to whether a shift in autonomy level occurred and, if so, in which direction (up or down). 
 
 In this work, we focus on information that derives from the human and their interaction with the autonomy, and thus on features computed from $\mathcal{H}$ and $\mathcal{I}$. Within a typical shared-control system, raw data streams from $\mathcal{H}$, $\mathcal{C}$, and $\mathcal{E}$ are available. Less typical are sensors that directly observe $\mathcal{I}$. Some measures of interaction however can be computed from $\mathcal{H}$ and $\mathcal{C}$ data streams, as is done in our implementation below. We use information from stream $\mathcal{E}$ as a point of comparison, to gauge how much events in the external environment, to which the human-robot team are responding, are captured within the human and interaction information streams.

\subsection{Implementation}
Three discrete Levels of Autonomy are presented for this study: 

\subsubsection{\textbf{$A_0$ - Teleoperation}} In $A_0$, the executed command directly maps the participant's input to the command output and no assistance is provided by the autonomy. 

\subsubsection{\textbf{$A_1$ - Autonomous Stopping}} In $A_1$, the autonomy prevents the participant from running into nearby obstacles by blocking commands that will lead to collisions and allowing direct command mapping otherwise. 

\subsubsection{\textbf{$A_2$ - Blended Autonomy}} In $A_2$, the wheelchair blends the human commands ($\textbf{u}_{h}$) with the autonomous obstacle avoidance planner ($\textbf{u}_{r}$) using a linear blending paradigm (\ref{eq:blending}) to produce an executed command, $\textbf{u}_{c}$. 
\begin{equation}
    \label{eq:blending}
    \textbf{u}_{c} = (1-\alpha) \cdot \textbf{u}_{r} + \alpha \cdot \textbf{u}_{h}
\end{equation} The blending parameter ($\alpha = 0.5$) is selected empirically for this study and is informed by prior work \cite{achmet1} that evaluated control paradigm $A_2$.

In our implementation, ground truth shifts in autonomy level derive from two sources: (1) the human operator, and (2) a safety-only automated shifting paradigm. Specifically, at any time during a trial, the participant can verbally request a shift up or down in assistance and the experiment proctor executes the shift. In our formulation, the human request to shift \textit{always supersedes} the autonomy request. Shift requests by the human or the autonomy beyond the minimum or maximum LOA are ignored. Audible alerts inform the participant of a successful shift in all instances.

The safety-only automated shifting paradigm operates as: 

\begin{equation}
    \label{eq:shifting}
\text {Shift } (d,t) = 
\begin{cases}
    \text{Up}, & d < \delta_l \text{ and } t > \tau \\
    \text{Down}, & d > \delta_h \text{ and } t > \tau 
\end{cases}
\end{equation} where $d$ is the distance to the nearest obstacle and $t$ is the time within (or outside of) the distance threshold. In our implementation, parameters $\delta_l = 0.65$, $\delta_h = 0.75$, and $\tau = 3$ seconds are set empirically. 

We include an automated shifting paradigm in our analysis in order to capture interactions between human-initiated and autonomy-initiated shifts in autonomy level. Of particular interest are scenarios where the human and autonomy \textit{disagree} about an LOA switch. We design our automated shifting paradigm to be simple and based on safety alone, in recognition of the fact that preserving the physical safety of the human-robot system is often the superseding objective of any autonomy system.

\section{EXPERIMENTAL METHODS AND DESIGN}
This section describes the experimental design and procedures used in this experiment.

\begin{figure}[t]
\vspace{0.5em}
\centering
\includegraphics[width=0.45\textwidth]{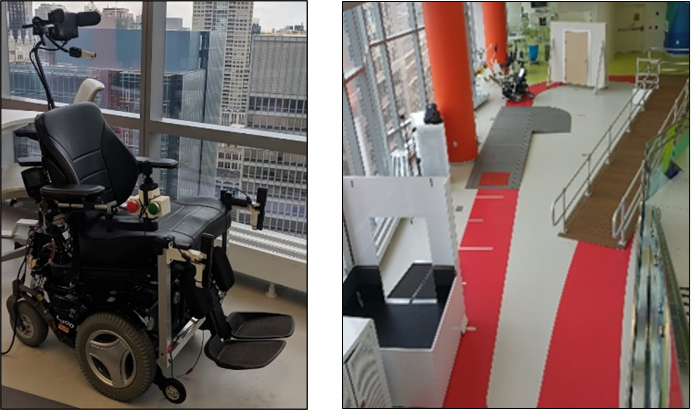}
\caption{The argallab autonomous wheelchair (left) and an example obstacle course (right).}
\label{fig:chairandcourse}
\end{figure}

\subsection{Hardware}
A customized, powered wheelchair is used for this experiment. The wheelchair, shown in Figure \ref{fig:chairandcourse}, is a commercially available (Permobil, Sweden) wheelchair fitted with a LiDAR, encoders, and an on-board computer. A software suite realized in ROS provides autonomy control commands. Human control inputs are provided via a 2-axis joystick. Section III.B describes the spectrum of shared-control paradigms that make use of both control signals.

\subsection{Procedure}
To complete the experimental tasks, the participants drive the robotic wheelchair through unique obstacle courses while completing a distraction task after undergoing a brief training period. The Northwestern University Institutional Review Board approved the experimental protocol and consent form. 

\subsubsection{Participants} The group consisted of 8 female and 8 male participants without motor impairments aged 21 to 37 years with varying levels of familiarity with robotic devices and wheelchairs. 

\subsubsection{Primary Task} Six distinct wheelchair obstacle courses are designed as different permutations of five navigation tasks: traversal of (1) a doorway, (2) a sidewalk ramp, (3) a narrow straight-away, (4) a tight turn, and (5) avoidance of a dynamic obstacle. Figure \ref{fig:chairandcourse} displays one such course. Participants complete a trial after they have returned to the specific course's starting location in the correct orientation. The LOA is initialized at the start of each trial, and then evolves according to human- and autonomy-initiated LOA shifts. Initial LOA assignment is counterbalanced according to a within-subjects design. The human operator is notified of changes in LOA via auditory alerts.

\begin{figure}[b]
\vspace{-1.25em}
\centering
\includegraphics[width=0.19\textwidth]{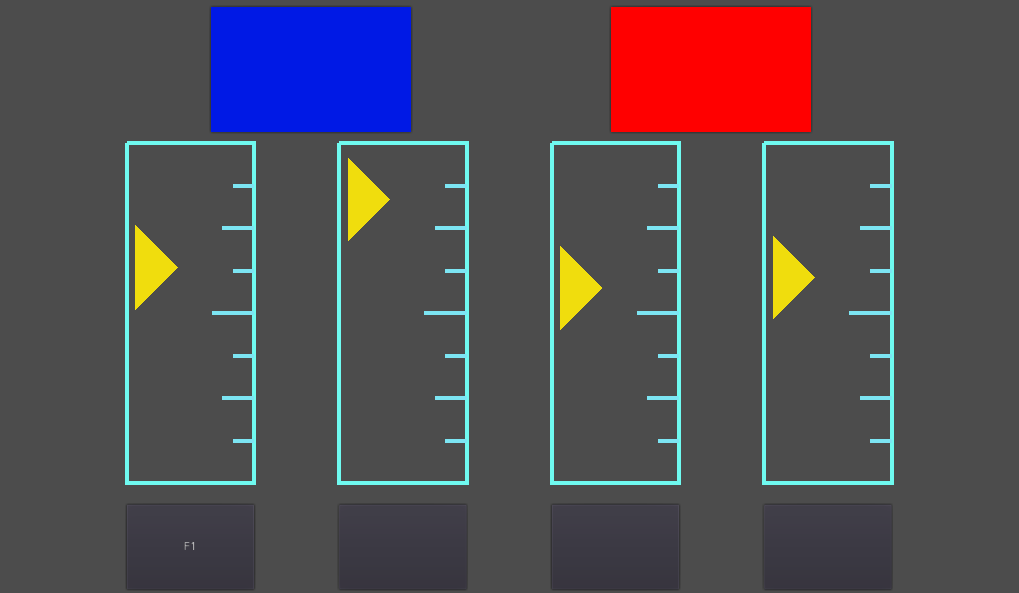}
\caption{The modified NASA MATB-II distraction task.}
\label{fig:dtask}
\end{figure}

\subsubsection{Secondary Task}
A wheelchair-mounted tablet PC displays a secondary ``distraction" task to the participants while they navigate the wheelchair through the obstacle course. This distraction task, shown in Figure \ref{fig:dtask}, increases the participant's cognitive load and is based on the ``gauges" subtask of the NASA Multi-Attribute Task Battery (MATB) \cite{NSATMATB}. Our implementation uses the U.S. Air Force Research Laboratory's suggested settings \cite{usafmatb}. In our experiment, the purpose of the distraction task is to \textit{modulate} the amount of cognitive load (rather than \textit{measure} load) as an additional means of modulating the complexity of the obstacle courses.  

To incentivize interaction with the secondary task, participants are required to respond to at least 75\% of malfunctions accurately or the experiment is paused, and they are unable to move the wheelchair until performance once again exceeds 75\% \cite{LOAsalloc}. Audible alerts inform participants to distraction task performance pauses. During the trials, no stopping occurred as all participants maintained $>75\%$ performance. 

\subsubsection{Participant Evaluation Metrics}
To evaluate perceived difficulty the raw, modified NASA-TLX, \textit{as presented in \cite{modTLX}}, is administered, which is quicker to administer than, and a comparable substitute to, the original NASA-TLX measure of perceived difficulty \cite{realTLX}. The modified NASA-TLX does not require pair-wise comparisons, but rather a 7-point Likert survey per NASA-TLX category. A higher score indicates higher perceived difficulty. 

\subsubsection{Protocol} At the start of the study session, the participant is given a description of the hardware and the course. LOA training consists of one traversal of a demonstration course, for each LOA. Distraction task training consists of interacting solely with the distraction task for two minutes.

At the start of each trial, the wheelchair is driven to the starting location for the set obstacle course. The proctor demonstrates the course's path by walking through it. The participant may ask for directions at any time. The participant is informed of their initial LOA and the trial is started. 

After completing each trial, the participant completes a modified version \cite{modTLX} of the NASA TLX survey.

\section{DATA PREPARATION AND LEARNING}

The following section describes the study's dataset, the extracted features, data labeling, and learning methods. 

\subsection{Signal Selections and Descriptions}
 We condition all model learning on the LOA in use when the data is gathered. The current LOA is always known to the autonomy system, and we expect characteristics of the computed features to vary with the LOA---for example, as the LOA increases the frequency of commands provided by the user decreases. The raw data are grouped into one of three information streams as presented in Table \ref{tab:sigs}. 

\begin{table}[ht]
\vspace{0.55em}
\begin{tabular}{p{3.5cm}|p{4cm}}
  \textbf{Information Stream} & \textbf{Raw Signal Description} \\
  
  \hline
  
  {Human Control Signals} ($\mathcal{H}$)  & \textbf{$j_{x}$, $j_y$}: The operator joystick commands along the joystick's x- and y-axes. \newline
   $\textbf{u}_H$ = $<$\textbf{$\omega_h$, $v_h$}$>$: The human rotational and linear velocity commands after translation from x-y joystick to linear-rotational values.\\
  \hline
  
  {Autonomy Signals}  ($\mathcal{C}$) & 
  
   $\textbf{u}_R$ = $<$\textbf{$\omega_r$, $v_r$}$>$: The autonomy rotational and linear velocity from the planner.\\
  
  \hline
  
  {Environment Signals}  ($\mathcal{E}$) & $d$: The raw depth data from the LiDAR.\\
  
\hline

\end{tabular}
\caption{INFORMATION STREAMS AND THEIR DATA SOURCE.}
\label{tab:sigs}
\vspace{-2em}
\end{table}

 This dataset is found to be highly imbalanced. To address the class imbalance, the over-represented class (no shift, 99.5\%) is randomly down-sampled to create a 90\%/10\% over-represented (no shift) / under-represented (shift-up and shift-down) class balance. Additionally, for deep learning methods, the minority classes are randomly up-sampled. The data are normalized and divided into train/validation/test sets following a 60/20/20 split. A total of 1044 shifts were requested by both participants and the naïve shifting algorithm.  On average, each trial had 10.8  $ \pm $ 4.38  shifts.

\subsection{Feature Extraction}
Five hand-designed features are selected to train on the classical machine learning methods: distance to nearest obstacle, smoothness of human control signals, frequency of human input, complex human-robot agreement, and simple human-robot agreement. 

\subsubsection{\textbf{Distance to Nearest Obstacle}} The autonomy identifies the nearest obstacle in the autonomy costmap and returns the obstacle's distance. Distance values are thresholded: $ \mathcal{E}_{d} \in [0.5 ,3]$ (\textit{meters}) to filter outliers that exist outside of the LiDAR's and autonomy performance specifications. 

\subsubsection{\textbf{Smoothness of Human Control Signals}} The smoothness of the human's input is measured using SPARC \cite{SPARC}, or Spectral Arc Length. This measure is selected for both its effectiveness at measuring smoothness among typically- and differently-abled populations as well as its strong robustness to noise.

\begin{equation}
    \label{eq:sparc1}
   \mathcal{H}_{\mathcal{S}} \gets - \int\limits_0 ^{\omega_c} \left[\left(\frac{1}{\omega_c}\right)^2+\left(\frac{d\hat{V}(\omega)}{d\omega}\right)^2\right]^{\frac{1}{2}} d\omega
\end{equation}
\begin{equation}
    \label{eq:sparc2}
   \hat{V}(\omega) = \frac{V(\omega)}{V(0)}
\end{equation}
\begin{equation}
    \label{eq:sparc3}
    \omega_c = \min\{\omega_c^{max}, \min\{ \omega, \hat{V}(r) < \bar{V}\;          \forall  \;  r > \omega\}\}
\end{equation} Here, $V(\omega)$ is the Fourier magnitude spectrum of the user's joystick inputs over a 2 second sliding window of commands, $\hat{V}(\omega)$ is the normalized Fourier magnitude spectrum of the user's inputs, $\bar{V}(\omega)$ is the amplitude threshold for computing the dynamic cutoff frequency, $\omega_c$ is a dynamic cutoff frequency tuned for noise sensitivity, $\omega_c^{max}$ is the joystick's bandwidth, and $\mathcal{H}_{\mathcal{S}}$ is the human smoothness feature. 

\subsubsection{\textbf{Frequency of Human Input}} Researchers in automotive engineering \cite{vehiclefreq} have used frequency of human input to classify the need for autonomous assistance. Researchers in rehabilitation robotics \cite{mahdieharxiv} have applied the formulation in (\ref{eq:freq1}-\ref{eq:freq2}) to robotic arm task difficulty.

\begin{equation}
    \label{eq:freq2}
     \mathcal{H}_{\Omega_E} \gets \beta \cdot \Omega_t + (1-\beta) \cdot \Omega_{t-1}
\end{equation}
\begin{equation}
    \label{eq:freq1}
   \Omega_t  =  \frac{1}{N} \sum\limits_{k = t-N}^{t} \frac{1}{t_k - t_{k-1}}
\end{equation} Here, $\Omega_t$ is the present input rate and $\beta$ is the weighting on the most-up-to-date measurement in an Exponential Moving Average, and $\mathcal{H}_{\Omega}$ is the human frequency feature.

\subsubsection{\textbf{Complex Human-Robot Agreement}} Comparing the human and the autonomy intentions, or measuring how much they ``agree," may indicate the need for an LOA change. Broad \textit{et al.} \cite{alexfrechet} compares a projection of the robot pose, based on the human's current command, to a projection of the robot pose, from the autonomy planner \cite{alexfrechet} using the discretized Fr\'echet distance \cite{frechet} as shown in (\ref{eq:frechet1} - \ref{eq:frechet2}). A constant time horizon, as set by the planner, is used when taking the projection of the robot pose from the autonomy planner. An N-point moving average of Fr\'echet distances is computed to filter noise. 

\begin{equation}
    \label{eq:frechet2}
\mathcal{I}_C \gets \frac{1}{N} \sum\limits_{i = t-N}^{t}  \mathcal{F}(\textbf{u}_{H_i},\textbf{u}_{R_i})
\end{equation}
\begin{equation}
    \label{eq:frechet1}
   \mathcal{F} = \inf_{\lambda, \zeta} \max_{t \in [0,1]} \left\{ d\left( A(\lambda(t)), B(\zeta(t)) \right)  \right\} 
\end{equation} Here, $A, B$ are two curves, $\inf$ is the infimum (or greatest lower bound),  $d$ is a distance function, $\textbf{u}_H$ and $\textbf{u}_R$ the human and autonomy projections, respectively, and $\mathcal{I}_C$ is the complex agreement feature.

\subsubsection{\textbf{Simple Human-Robot Agreement}} Inspired by the Complex Human-Robot Agreement feature, a simpler feature compares only the instantaneous commands provided by the human and the autonomy at a given time step, ignoring possible future commands. Here, the L2-normed difference between the instantaneous human and autonomy commands is computed. To filter noise, an N-point moving average of these differences is also computed.
 \begin{equation}
    \label{eq:simple}
   \mathcal{I}_s \gets \frac{1}{N} \sum\limits_{i = t-N}^{t} ||\textbf{u}_{H_i} - \textbf{u}_{R_i}||_2
\end{equation} Here, $\textbf{u}_H$ and $\textbf{u}_R$ are the instantaneous human and autonomy commands, respectively, and $\mathcal{I}_s$ is the simple agreement feature.

\subsection{Classical Learners}
Five different classical machine learning classification algorithms were selected for training: support vector machines (SVM), random forests (RF), gradient boosting classifiers (GBC), logistic regression (LR), and Naïve Bayes (NB). These methods were selected as to span a breadth of classical machine learning techniques. \verb|scikit-learn| was used to implement all classifiers and performance metrics \cite{sklearn}. 

To test the performance of each information stream in label prediction, models were trained using different datasets containing only certain features as shown in Table \ref{table:classicalstreams}. 

\begin{table}[t]
\vspace{0.55em}
\centering
\begin{tabular}{l|l}

\multicolumn{1}{l|}{\textbf{Information Stream}} & \textbf{Features Analyzed} \\ \hline
$\mathcal{H}$ &  $(\mathcal{H}_{\Omega},  \mathcal{H}_{\mathcal{S}}) $ \\ \hline
$\mathcal{I}$ &$(\mathcal{I}_{C},  \mathcal{I}_{\mathcal{S}})$  \\ \hline
$\mathcal{H}$ + $\mathcal{I}$ + $\mathcal{E}$  &$(\mathcal{H}_{\Omega}, \mathcal{H}_{\mathcal{S}}, 
\mathcal{I}_{C},  \mathcal{I}_{\mathcal{S}}, \mathcal{E}_{d} )$                            \\ \hline
\end{tabular}
\caption{HAND-ENGINEERED FEATURES COMPARED USING CLASSICAL LEARNING.}
\label{table:classicalstreams}

\centering
\begin{tabular}{l|l}
\multicolumn{1}{l|}{\textbf{Information Stream}} & \textbf{Raw Signals Analyzed} \\ \hline
$\mathcal{H}$ &  $(j_x, j_y, \omega_h, v_h)$ \\ \hline
$\mathcal{I}$ & $(\omega_r, \omega_h, v_r , v_h)$  \\ \hline
$\mathcal{H}$ + $\mathcal{I}$ + $\mathcal{E}$  & $(\omega_r, \omega_h, v_r , v_h, J_x, J_y, d)$                            \\ \hline
\end{tabular}
\caption{RAW SIGNALS COMPARED USING DEEP LEARNING.}
\label{table:deepstreams}
\vspace{-3em}
\end{table}

To select the best hyperparameters for each learner, a grid search optimizing over macro-$F_1$ scores \cite{MacroF1} was performed for each learner, at each LOA, and each (group of) information stream(s). The grid search was realized using the \verb|hypopt| package which allows for the use of a specified validation dataset and parallel learner training. In total 9,072 models were trained.

\subsection{Deep Learning}
Six different deep learning models were selected for this task. Each model was inspired by shared control literature or similarly structured research problems. Table \ref{table:deepmodels} presents each model's components. All models were trained on the class-balanced datasets. Models were validated and tested using the non-class-balanced data. The best performing model within each LOA was selected by comparing balanced accuracy scores. All models underwent early-stopping when the model's validation loss increased or remained constant for five epochs. 120 models were trained.

\begin{table}[ht]
\vspace{0.55em}
\begin{tabular}{p{3.5cm}|p{4cm}}
  \textbf{Model Name} & \textbf{Model Description} \\
  
  \hline
  \verb|2DCNN| & \textbf{3x 2D CNNs}: \textit{filters} (in order): 32, 64, and 128,                         \textit{kernel}: (2,2)\newline
                \textbf{MaxPool2D}: \textit{size}: 2 \newline  
                \textbf{Dropout}: \textit{rate}: 0.10 \newline  
                \textbf{3x Dense Layers}: \textit{Number of Units} (in order): 500, 250, 50\\
 \hline
 \verb|LSTM Only| & \textbf{3x LSTM Units}: \textit{Number of Units}: 200 each\newline                  
                 \textbf{3x Dense Layers}: \textit{Number of Units} (in order): 500, 250, 50\\
\hline
\verb|BLSTM Only| & \textbf{3x Bi-Directional LSTM Units}: \textit{Number of Units}: 200 each\newline
                 \textbf{3x Dense Layers}: \textit{Number of Units} (in order): 500, 250, 50\\
\hline
\verb|ConvLSTM2D| & \textbf{1x ConvLSTM2D}: \textit{filters}: 64 each,                                                       \textit{kernel}: (1,3)\newline
                    \textbf{Dropout}: \textit{rate}: 0.10 \newline
                    \textbf{1x Dense Layers}: \textit{Number of Units}: 100\\
 \hline
 \verb|1DCNN-LSTM| & \textbf{2x 1D CNNs}: \textit{filters}: 64 each,                                             \textit{kernel}: 5\newline
                    \textbf{Dropout}: \textit{rate}: 0.10 \newline
                    \textbf{MaxPool1D}: \textit{size}: 4 \newline  
                    \textbf{1x LSTM Unit}: \textit{Number of Units}: 100 \newline
                    \textbf{1x Dense Layers}: \textit{Number of Units}: 100\\
 \hline
 \verb|2DCNN-LSTM| & \textbf{3x 2D CNNs}: \textit{filters} (in order): 32, 64, and 128,                                      \textit{kernel}: (2,2)\newline
                \textbf{MaxPool2D}: \textit{size}: (3,3) \newline  
                \textbf{Dropout}: \textit{rate}: 0.10 \newline 
                \textbf{3x LSTM Units}: \textit{Number of Units}: 200 each\newline
                \textbf{3x Dense Layers}: \textit{Number of Units} (in order): 500, 250, 50\\
 \hline
\end{tabular}
\caption{DEEP LEARNING MODELS TRAINED AND THEIR ARCHITECTURES.}
\label{table:deepmodels}
\vspace{-2.8em}

\end{table}

Table \ref{table:deepstreams} presents the signals used in each information stream for deep learning. Note that in our implementation, no raw interaction data stream $\mathcal{I}$ exists. In this category, we provide to our deep learners the same raw data streams used to compute the hand-engineered interaction features $\mathcal{I}$.

\section{RESULTS}
The results indicate human-only data streams do indeed carry information relevant for LOA switching. The interaction data streams are found to carry even more relevant information, when properly extracted, and there is a suggestion of significant overlap in the information encoded within the interaction and environment data streams. The results also provide interesting insights into the availability of shift-relevant information within these recorded data streams.

To compare information stream classification performance, aggregate balanced accuracy is computed. This score computes a weighted accuracy, where the arithmetic mean of the class-wise accuracies for each LOA-specific model is weighted by the proportion of class labels present in the training dataset. This score treats each class with equal importance, and was used in lieu of the more typical (balanced) accuracy to address the strong class imbalance. As a point of comparison, an aggregate chance score is computed from the weighted average of each LOA's individual chance scores. 

\subsection{Classical Learning Results}

\begin{figure}[b]
\centering
\vspace{-1em}
\includegraphics[width=8.5cm]{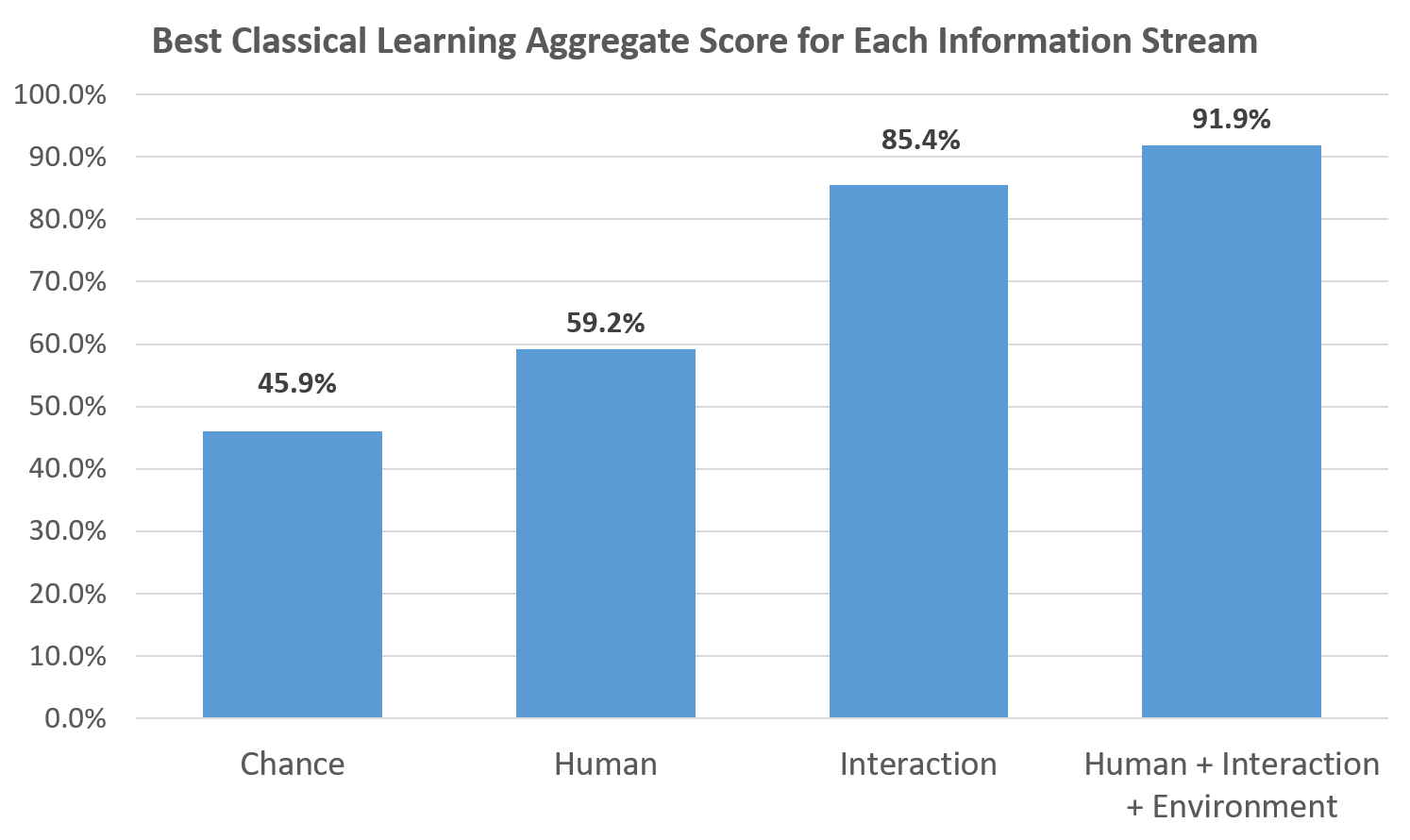}
\vspace{-1em}
\caption{The best classical learner per information stream's aggregate balanced accuracy on the test data.}
\label{plot:best_class_bacc}
\end{figure}

To compare the predictive power of each of the individual and combined information streams when using hand-engineered data features, the best classical learners for each stream are reported in Table \ref{table:thetable} and their aggregate accuracies performance plotted in Figure \ref{plot:best_class_bacc}.

Monitoring only $\mathcal{H}$ encodes sufficient information regarding when to shift LOA to perform markedly better than chance (59.2\% vs. 45.9\%). Significantly more information appears to be encoded within the interaction between the human and the autonomy (85.4\% vs. 59.2\%). As a point of comparison, considering all information streams ($\mathcal{H}$ + $\mathcal{I}$ + $\mathcal{E}$) provides only a modest improvement over $\mathcal{I}$ alone (91.9\% vs. 85.4\%).

Data streams differ in terms of the computational cost to extract meaningful information from them and in terms of the richness of the resulting information. Often the two are in lock step, where richer information comes at the cost of greater computation. For example, processing environment data as derived from a large point cloud is more costly than computing features from a joystick commands. In light of this, the fact that data from the environment only encodes a comparatively small additional amount of information ($\sim$5\% increase in classification accuracy) relevant to when to shift LOA, in comparison to measures of difference (disagreement) between the human and autonomy control commands, is a particularly interesting result that merits further investigation in future work.

\begin{table}[t]
\vspace{0.55em}
\centering
\begin{tabular}{p{1.4cm}|p{2.5cm}|p{2.7cm}}

\textbf{Information} \newline 
 \textbf{Stream} &
 \textbf{Best Classical}\newline 
 \textbf{Learner}\newline
 \textbf{(Agg. Error)} & 
 \textbf{Best Deep}\newline
 \textbf{Learner}\newline 
 \textbf{(Agg. Error)} \\ 

\hline
$\mathcal{H}$                                 &  RF (40.8\%)   & \verb|ConvLSTM2D| (40.5\%) \\ \hline
$\mathcal{I}$                                 &  GBC (14.6\%)   & \verb|2DCNN| (41.4\%) \\ \hline
$\mathcal{H}$ + $\mathcal{I}$ + $\mathcal{E}$ &  GBC (8.1\%)   & \verb|LSTM Only| (14.0\%)   \\ \hline
\end{tabular}
\caption{BEST CLASSICAL AND DEEP LEARNER PERFORMANCE (AGGREGATE ERROR) FOR EACH INFORMATION STREAM.}
\label{table:thetable}
\vspace{-3em}

\end{table}

\subsection{Deep Learning Results}
\begin{figure}[b]
\vspace{-2em}
\includegraphics[width=8.5cm]{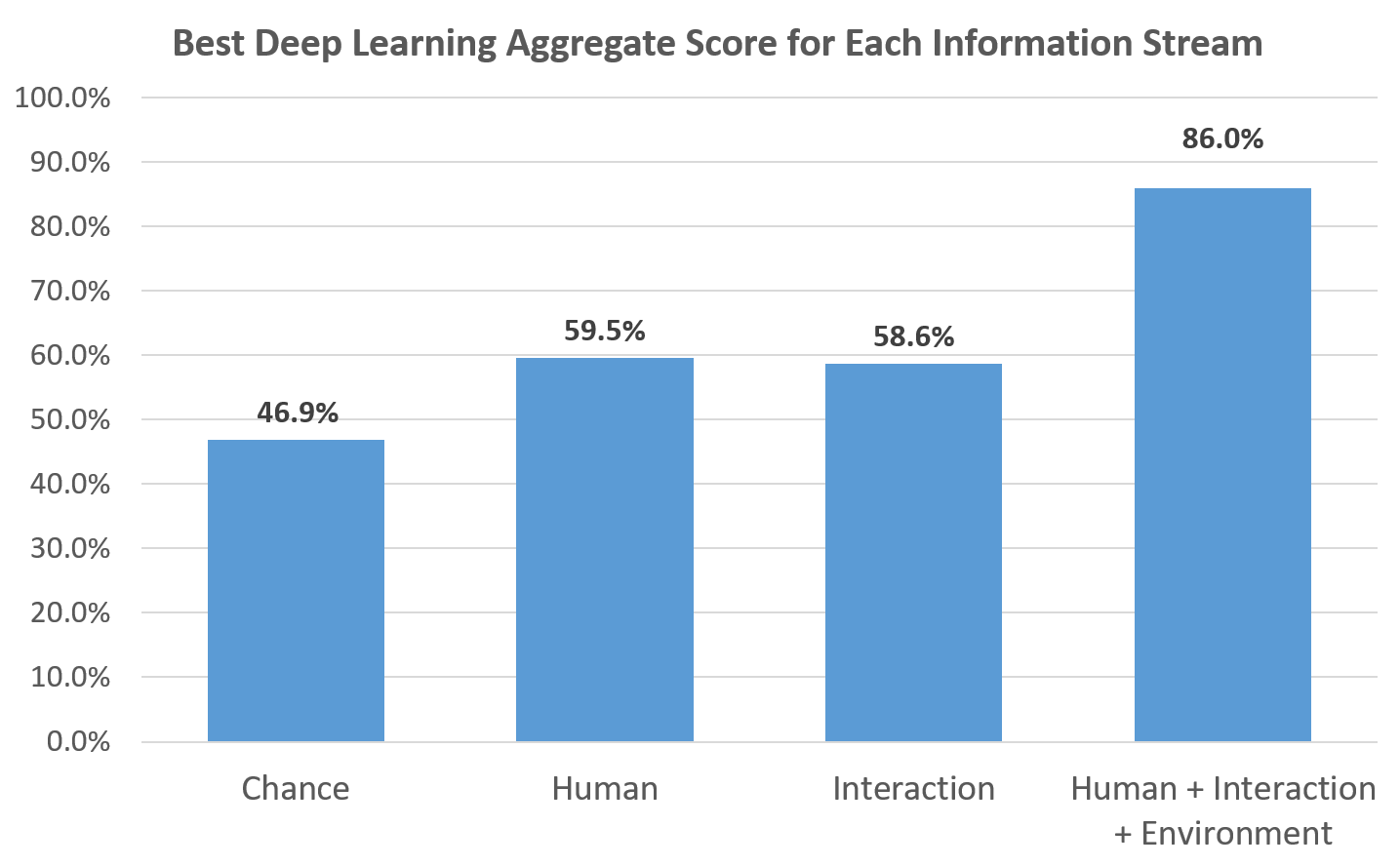}
\vspace{-1em}
\caption{The best deep learner per information stream's aggregate balanced accuracy on the test data.}
\label{plot:best_deep_bacc}
\end{figure}

  To compare the predictive power of each information stream when using learned features, the best deep learners for each stream are reported in Table \ref{table:thetable} and their aggregate accuracies performance plotted in Figure \ref{plot:best_deep_bacc}. 

The results show equivalent performance to the classical learners when using raw data stream $\mathcal{H}$. Thus, the deep networks are able to learn the hand-engineered $\mathcal{H}$ features of smoothness and frequency; or rather, they are able to learn some representation that encodes an equivalent amount of information regarding when to shift LOA. Notably, they do not uncover a representation that encodes more information about when to shift LOA. 

A marked performance drop is observed when using raw data stream $\mathcal{I}$ (58.6\% vs. 85.4\%). This suggests that the hand-designed interaction features ($\mathcal{I}$) of agreement were not able to be learned, and neither did the deep network learn a representation that encodes a similar amount of information regarding when to shift LOA.

Interestingly, when provided with all data streams ($\mathcal{H}$ + $\mathcal{I}$ + $\mathcal{E}$), the deep learners' performance is on par with the classical learners interaction-only ($\mathcal{I}$) performance (86.0\% vs. 85.4\%). This result suggests an overlap in the shift-relevant information encoded within the $\mathcal{I}$ and $\mathcal{E}$ information streams, which is further supported by the observation that the inclusion of environment information only provides a comparatively modest improvement in the classical learner performance (91.9\% vs. 85.4\%).

 It is worthwhile to note that many deep networks struggle to achieve good performance with highly imbalanced datasets, even after data augmentation \cite{deepnets_imbalance}. Certain hand-designed features (e.g., SPARC) may include operations, such as the square root function, that are difficult to estimate using common feature extraction methods used in deep learning. The classical methods use hand-designed features, while the deep nets learn these features. The difference in performance between classical learners and deep learners can in part be attributed to this difference in hand-designed versus learned features. Each individual method, whether classical or deep learning, also differs in the structure of the function that is learned. Lastly, we note that the environment information stream is used, raw, by the safety-only automated shifting paradigm, and so provides shift-relevant information without the need for feature extraction beyond simple thresholding. Thus, one possible explanation for the significant improvement in deep learner performance with the addition of environment information is that the relation between this information stream and LOA shifts was straightforward to extract and accordingly easier for the network to understand.

\subsection{Analysis of Perceived Human Difficulty}
The modified NASA TLX surveys range from 0 to 42 where a higher score indicates higher perceived difficulty. The administered surveys indicates a similar level of average (over trials) perceived difficultly across all participants ($28.9 \pm 2.5$). Similarly, the survey data demonstrates little variance in perceived course difficulty when averaged across all participants for a given course ($28.9 \pm 1.0$).

\section{CONCLUSIONS}

This work seeks to inform the design of dynamic autonomy allocation frameworks by analyzing the classification power of different information streams within the human-robot team. The results indicate that human-only information streams alone encode sufficient information to measurably out-perform chance, demonstrating the potential for autonomy shifting frameworks that rely on human inputs alone. However, monitoring signals related to the \textit{interaction} between the autonomy and the human within a human-robot team, when properly encoded, yields the most dramatic jump in classification accuracy. The features learned by deep networks prove variable in their ability to represent shift-relevant information, in comparison to the hand-engineered features. The suggestion that some environment information also is encoded within the interaction data is an interesting observation that will be further pursued in future work. 

 \addtolength{\textheight}{-4cm}   



\section*{ACKNOWLEDGMENT}
 We would like to gratefully acknowledge support from U.S. Office of Naval Research under the Award Number N00014-16-1-2247. The research reported in this publication also was supported by a grant from the U.S. Department of Defense through the National Defense Science \& Engineering Graduate (NDSEG) Fellowship Program under the fellowship number F-6826820873, as well as by the National Institute of Biomedical Imaging and Bioengineering of the National Institutes of Health under Award Number R01-EB024058. The content is solely the responsibility of the authors and does not necessarily represent the official views of the National Institutes of Health. 
 
 The authors thank Alisha Rege, Ravender Virk, and Dave Witson for their data engineering suggestions and machine learning implementation insights.

\bibliographystyle{IEEEtran} 
\bibliography{workscited.bib}

\end{document}